\documentclass{article}

\PassOptionsToPackage{numbers, compress}{natbib}

\usepackage{amsmath}
\usepackage{multirow}
\usepackage{graphicx}
\usepackage{wrapfig}
\usepackage{color}

\usepackage[preprint]{neurips_2024}

\usepackage{wrapfig}
\usepackage[utf8]{inputenc} 
\usepackage[T1]{fontenc}    
\usepackage{hyperref}       
\usepackage{url}            
\usepackage{booktabs}       
\usepackage{amsfonts}       
\usepackage{nicefrac}       
\usepackage{microtype}      
\usepackage{xcolor}         

\title{OmniBind: Large-scale Omni Multimodal Representation via Binding Spaces}

\newcommand*{\affaddr}[1]{#1} 
\newcommand*{\affmark}[1][*]{\textsuperscript{#1}}

\author{
\textbf{Zehan Wang}\affmark[1]\thanks{Equal contribution. \{wangzehan01, ziangzhang\}@zju.edu.cn} \ \ \ \  \textbf{Ziang Zhang}\affmark[1]\footnotemark[1]\ \ \ \  \textbf{Hang Zhang}\affmark[1]\ \ \ \  \textbf{Luping Liu}\affmark[1]\ \ \ \  \textbf{Rongjie Huang}\affmark[1]\\ \ \ \  \textbf{Xize Cheng}\affmark[1]\ \ \ \  \textbf{Hengshuang Zhao}\affmark[2]\ \ \ \  \textbf{Zhou Zhao}\affmark[1]\\
\affaddr{\affmark[1]Zhejiang University\ \ \ }
\affaddr{\affmark[2]The University of HongKong\\}
\url{http://omnibind.github.io}
}

\begin{document}

\maketitle

\begin{figure}[h]
    \vspace{-2\baselineskip}
    \centering
    \includegraphics[width=1\linewidth]{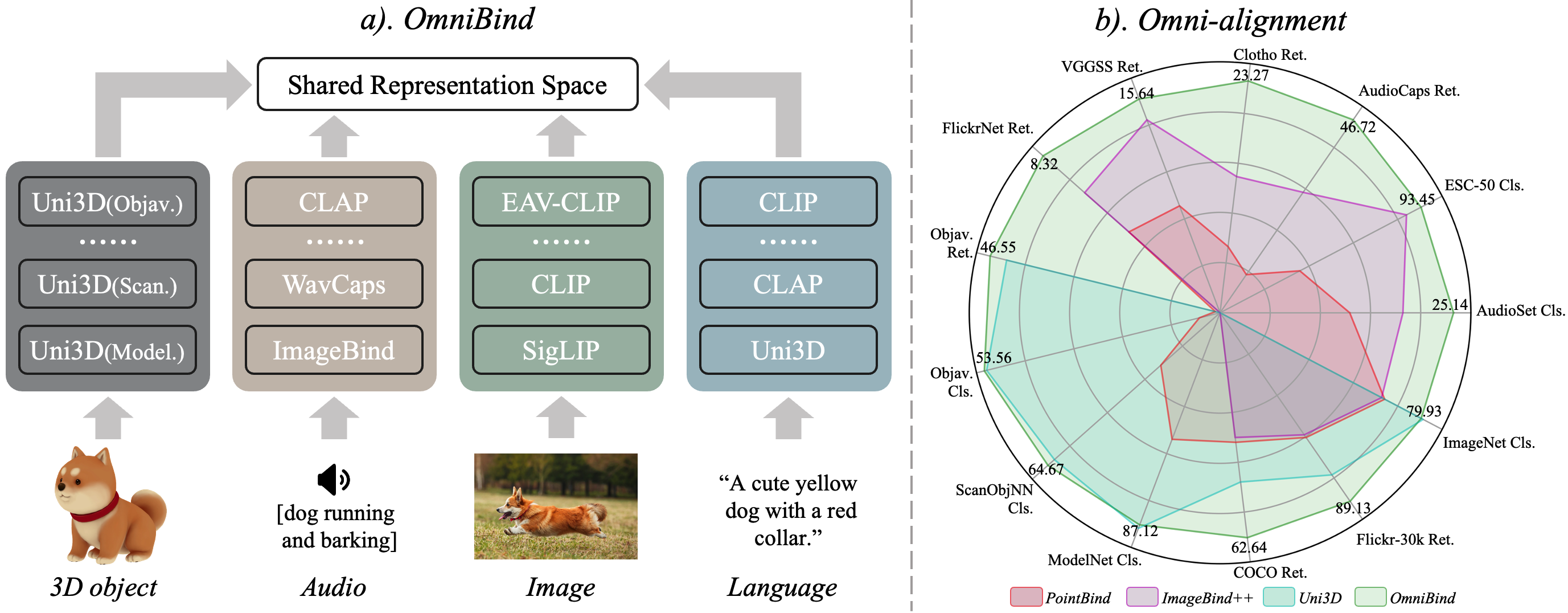}
    \caption{\textbf{Overview of OmniBind.} OmniBind integrates diverse knowledge of various existing multimodal models, leading to large-scale omni representations. OmniBind exhibits remarkable versatility and achieves state-of-the-art results on extensive downstream tasks over all modality pairs.}
    \label{fig:view}
\end{figure}
\begin{abstract}
Recently, human-computer interaction with various modalities has shown promising applications, like GPT-4o and Gemini. Given the foundational role of multimodal joint representation in understanding and generation pipelines, high-quality omni joint representations would be a step toward co-processing more diverse multimodal information. In this work, we present \textbf{OmniBind}, large-scale multimodal joint representation models ranging in scale from 7 billion to 30 billion parameters, which support 3D, audio, image, and language inputs.
Due to the scarcity of data pairs across all modalities, instead of training large models from scratch, we propose remapping and binding the spaces of various pre-trained specialist models together. This approach enables "scaling up" by indirectly increasing the model parameters and the amount of seen data. To effectively integrate various spaces, we dynamically assign weights to different spaces by learning routers with two objectives: cross-modal overall alignment and language representation decoupling. Notably, since binding and routing spaces both only require lightweight networks, OmniBind is extremely training-efficient. Learning the largest 30B model requires merely unpaired unimodal data and approximately 3 days on a single 8-4090 node. Extensive experiments demonstrate the versatility and superiority of OmniBind as an omni representation model, highlighting its great potential for diverse applications, such as any-query and composable multimodal understanding.
\end{abstract}

\section{Introduction}
Multimodal joint representation, which aligns different modalities into a shared space, forms the foundation of current multimodal understanding~\cite{liu2024visual, liu2023improved, bai2023qwen, zhu2023minigpt} and generation pipelines~\cite{rombach2022high, podell2023sdxl,  singer2022make, huang2023make}. Recently, co-understanding and generating various modalities with omni model has attracted increasing attention and demonstrates promising application prospects, like GPT-4o~\cite{gpt4o} and Gemini~\cite{team2023gemini}. However, existing open-source multimodal representation models primarily explore on a relatively small scale and are trained for limited combinations of modalities, such as image-text~\cite{radford2021learning, zhai2023sigmoid, sun2024eva}, audio-text~\cite{laionclap2023, mei2023wavcaps}, audio-image~\cite{gong2022contrastive, girdhar2023imagebind} and 3D-image-text~\cite{xue2023ulip, xue2023ulip2, liu2024openshape, zhou2023uni3d}.


Scaling up achieves incredibly success in language~\cite{achiam2023gpt, touvron2023llama, jiang2024mixtral, touvron2023llama2} and vision-language models~\cite{alayrac2022flamingo, liu2023improved, sun2024eva}. It consistently improves model performance and generalization by increasing the seen data and model parameters. Therefore, it is appealing to develop a large-scale omni multimodal representation model. However, due to the scarcity of paired data and scalable architectures across every modality, it is difficult to train omni multimodal representation models from scratch.

In this paper, we present \textbf{OmniBind}, large-scale unified representation models incorporating 3D, audio, image, and language. Instead of training large models from scratch, we bind numerous spaces together, thereby developing an ensemble model that has indirectly seen massive data and contains a large number of parameters, as shown in Fig.~\ref{fig:view}-\textit{a}. By integrating the alignment knowledge between different modalities, the resulting model pieces together omni alignment across all modalities. Moreover, inheriting pre-trained knowledge greatly reduces the computing resources and data requirements. The entire learning process can be completed with several million unpaired data and 4090 GPUs.


However, it is non-trivial to effectively integrate numerous pre-trained spaces. Simply remapping and weighted-averaging spaces as ~\cite{wang2024freebind}, cannot achieve the desired "scaling up". As more spaces are integrated, knowledge from different sources would interfere with each other. Manually adjusting the combining factors only results in trade-offs between different expertise rather than a truly versatile omni model. We attribute the interference and trade-off phenomenon to the fixed combining weight. Spaces trained for different purposes on varied datasets contain knowledge of various aspects, and averaging them with fixed weights can only preserve knowledge of certain aspects.



Inspired by the Mixture-of-Expert technique in LLMs~\cite{fedus2022switch, jiang2024mixtral, lepikhin2020gshard, du2022glam}, we introduce weights routing strategy to ensemble the spaces, aiming to resolve the above dilemma. Specifically, for each modality, we utilize a learnable router to dynamically predict the combining weights based on input information. Two main objectives guide the training of routers: cross-modal overall alignment and language representation decoupling. The former motivates routers to predict optimal weights for all modality combinations, while the latter is designed to alleviate the conflicts between text embeddings that are aligned to different modalities, and ensure the discrimination of language representations.

With above techniques, we bind 14 existing spaces together, successfully developing three omni models ranging from 7 to 30 billion parameters, named \textbf{OmniBind}. Our method showcases impressive performance on extensive multimodal tasks. For 3D, audio, and image classification, OmniBind shows advanced zero-shot generalization ability. It also demonstrates significantly stronger cross-modal alignment on all possible modality pairs. In addition, benefiting from the high-quality omni semantic alignment, OmniBind enables impressive applications, such as accurate 3D-audio retrieval, any-query object localization/audio separation, and complex composable understanding.

Our contributions can be summarized as follows:

1) We propose \textbf{OmniBind}, large-scale omni multimodal representation models, ranging in scale from 7 billion to 30 billion parameters and encompassing four modalities: 3D point, audio, image, and language. It emphasizes the value of piecing various pre-trained specialist models together.

2) We introduce routers to ensemble spaces pre-trained on various modalities and datasets, thereby mitigating interference between knowledge from different sources and further enhancing versatility.

3) We design two learning objectives for learning routers: cross-modal overall alignment and language representation decoupling, which motivate routers to dynamically predict the optimal combining weights for all modality pairs while reserving the discrimination of representations.

4) OmniBind exhibits state-of-the-art performance on 13 benchmarks that cover all the modality pairs, and great potential for diverse applications, such as 3D-audio retrieval and any-query separation/localization, while requiring minimal computing resources and data.


\section{Related Work}
\subsection{Multimodal Joint Representation} Multimodal joint representation mainly aims to map inputs from different modalities into a shared space. Leveraging the semantic alignment property, pre-trained representation models are widely utilized in current multimodal large language models, such as LLaVA~\cite{liu2024visual, liu2023improved}, ImageBind-LLM~\cite{han2023imagebind}, and Chat-3D~\cite{wang2023chat, huang2023chat}, as well as multimodal generation pipelines like Stable Diffusion~\cite{rombach2022high}, SD XL~\cite{podell2023sdxl}, make-a-video~\cite{singer2022make} and make-an-audio~\cite{huang2023make}.

The initial multimodal representation is CLIP model~\cite{radford2021learning}, which demonstrates impressive generalization capabilities in various vision-language downstream tasks. Motivated by its success, successors propose stronger image-language representations by using higher-quality initialization~\cite{fang2023eva, sun2023eva}, larger-scale datasets~\cite{schuhmann2022laion, kakaobrain2022coyo-700m}, improved learning objectives~\cite{zhai2023sigmoid}, or better model architecture~\cite{fang2023eva2, li2023scaling}. Besides, some researchers employ the multimodal contrastive learning paradigm in other modalities pair. CLAP~\cite{laionclap2023} and WavCaps~\cite{mei2023wavcaps} learn aligned audio-language space from audio-text pairs.

In addition to these studies improving the alignment of two modalities, another line of work aims to develop unified spaces capable of accommodating inputs from multiple modalities (more than two). For instance, AudioCLIP~\cite{guzhov2022audioclip} and WAV2CLIP~\cite{wu2022wav2clip} introduce additional audio encoders for CLIP, leveraging audio-image-text and audio-image pairs. ULIP~\cite{xue2023ulip} and Uni3D~\cite{zhou2023uni3d} collect massive amounts of 3D-image-text paired data, enabling them to learn 3D-image-text joint representations based on advanced image-text pre-trained models.
More recently, ImageBind~\cite{girdhar2023imagebind} and LanguageBind~\cite{zhu2023languagebind} propose to integrate multiple modalities using different data pairs that share the crucial image or language modality. 

While current methods showcase a certain degree of robust cross-modal semantic alignment, they are primarily explored at relatively small scales and limited to specific modality pairs. On the other hand, our work focuses on large-scale omni representation models proficient in all modality pairs.

\subsection{Scaling Up Models}
Scaling up language or vision-language models has been incredibly successful in perception, reasoning, and generation tasks. GPT4~\cite{achiam2023gpt} and LLaMA~\cite{touvron2023llama, touvron2023llama2} achieve impressive conversation capability via training billion level language models on almost the massive internet language data. Recent InternVL~\cite{chen2023internvl} and EVA-CLIP-18B~\cite{sun2024eva} try to scale the parameter of the CLIP model to the ten-billion level. These methods prove that with the growth of model parameters and seen data, models obtain consistent and non-saturating performance improvement.

However, due to the scarcity of paired data across all modalities, the development of large-scale omni multimodal representation models has been lagging. In this paper, we propose to achieve the scaling up by binding existing spaces pre-trained on different modalities and datasets, rather than learning larger models from scratch with more data. This approach results in ensemble models that incorporate diverse knowledge from massive data and range from 7 billion to 30 billion parameters.


\subsection{Knowledge Fusion} 
Fusing knowledge from different sources is a classical and wildly-used method to develop robust AI models. Traditional ensemble learning methods~\cite{zhou2021ensemble, zounemat2021ensemble} train models with different sub-datasets, and combine the output of different models as the final prediction. Similar ideas are also employed by large language model research, recent works~\cite{bansal2024llm, wan2024knowledge, yu2023language} propose to merge multiple language models tuned for different downstream tasks, and the resulting model excels at all aspects. Moreover, the Mixture-of-Experts (MoE) language model~\cite{fedus2022switch, jiang2024mixtral, chowdhery2023palm} is also developing a hybrid model consisting of multiple sub-models and obtains better performance or efficiency by integrating them together.

For multimodal representation learning, C-MCR~\cite{wang2023connecting} and Ex-MCR~\cite{wang2023extending} first propose to fuse two bi-modality representation space via the shared modality, thereby building unified space with low data and computing resource requirements. FreeBind~\cite{wang2024freebind} is a high-level abstract of the above two methods, which employs bi-modality spaces to augment pre-trained unified space. Unlike these methods that only involve a few numbers of spaces, our goal is to obtain large-scale omni representations via binding extensive spaces, which face more severe risks of knowledge interference.

\begin{figure}[t]
	\centering
        \vspace{-1\baselineskip}
	\includegraphics[width=0.9\linewidth]{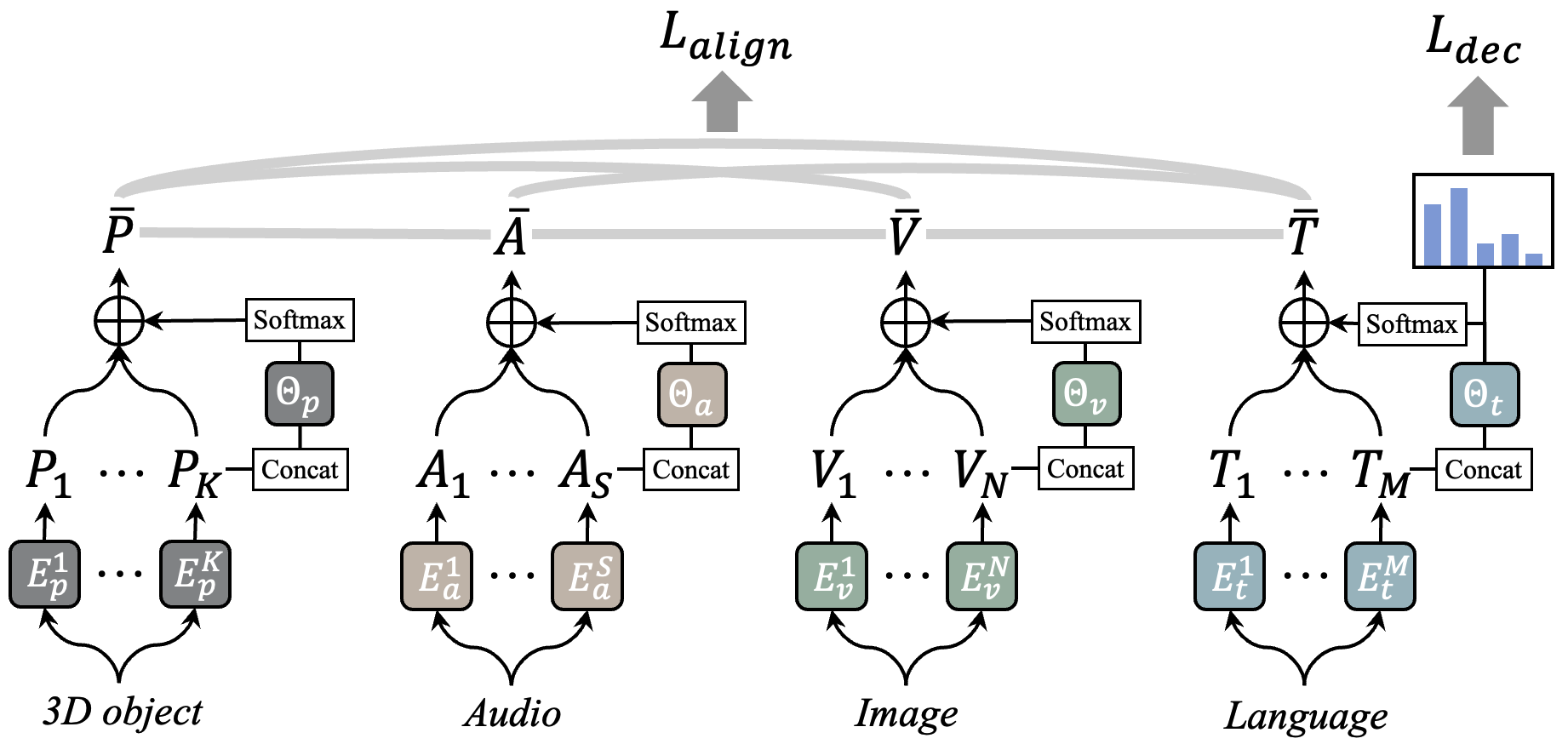}
	\caption{The pipeline of \textbf{OmniBind}. The $\Theta_{X}$ denotes the router of modality $X$, and $E_{X}^i$ is the $i$-th encoder of modality $X$. The losses $L_{align}$ and $L_{dec}$ are the objectives for training the routers.}
        \vspace{-1\baselineskip}
        \label{fig:model}
\end{figure}

\section{OmniBind}
\subsection{Spaces Binding}
\label{sec:binding}
Given that vision-language models are well-resourced and play a pivotal role in the multimodal field, we choose the advanced CLIP model EVA-CLIP-18B~\cite{sun2024eva} as the foundation, and bind additional image-text, audio-text, audio-image-text and 3D-image-text spaces onto it. 

For space binding, FreeBind~\cite{wang2024freebind} firstly proposes to improve a representation space by integrating extra spaces. Its space binding pipeline can be summarized as two steps: 1) collecting pseudo embedding pairs across two spaces and 2) mapping one space to another. Our binding process is primarily derived from FreeBind, but we replace its pseudo embedding-pair aggregation with a more efficient and robust pseudo item-pair retrieval, and scale up the number of integrated spaces.

Concretely, FreeBind first encodes massive unpaired unimodal data into embeddings of each space, and then uses the cross-modal similarity maps to aggregate pseudo embedding pairs across two spaces. The embedding pairs are unique for each pair of spaces. Therefore, when binding extensive spaces, repeatedly aggregating embeddings would be resource-intensive. Besides, the non-shared pseudo pairs are also unstable due to the varying performance of existing spaces.

To bind spaces robustly and efficiently, we directly retrieve pseudo pairs across all modalities at the item level.
Considering lots of unpaired 3D, audio, image, and language data, and leveraging the most advanced 3D-image-text, audio-text, audio-image, and image-text retrieval model, we can take each modality as a starting point to retrieve the top-1 recall of data from other modalities. This approach constructs the pseudo item pairs $\{p, a, v, t\}$. 

Using the pseudo data, we train simple projectors to individually bind each space to EVA-CLIP-18B. The training objective of the projector is the multimodal contrastive loss between all pairs of involved modalities. For instance, when binding CLAP with EVA-CLIP-18B, the learning objective is:
\begin{equation}
\label{eq:l_proj}
    L_{bind} = \operatorname{Info}(\Psi(\mathbf{A}_{at}), \mathbf{T}_{vt}) + \operatorname{Info}(\Psi(\mathbf{A}_{at}), \mathbf{V}_{vt}) + \operatorname{Info}(\Psi(\mathbf{T}_{at}), \mathbf{T}_{vt}) + \operatorname{Info}(\Psi(\mathbf{T}_{at}), \mathbf{V}_{vt})
\end{equation}
where $\mathbf{A}_{at}$, $\mathbf{T}_{at}$ are CLAP embeddings, $\mathbf{V}_{vt}$, $\mathbf{T}_{vt}$ are EVA-CLIP-18B embeddings and the $\Psi(\cdot)$ is an multi-layer perceptron projector. The $\operatorname{Info(\cdot,\cdot)}$ is multimodal constrastive loss:
\begin{equation}
    \operatorname{Info(\mathbf{X},\mathbf{Y})} = - \frac{1}{2} \sum^B_{i=1} (\operatorname{log}\frac{e^{(\mathbf{x_i} \cdot \mathbf{y_i}) / \tau}}{\sum^N_{j=1} e^{(\mathbf{x_i} \cdot \mathbf{y_j}) / \tau} } + \operatorname{log}\frac{e^{(\mathbf{y_i} \cdot \mathbf{x_i}) / \tau}}{\sum^N_{j=1} e^{(\mathbf{y_i} \cdot \mathbf{x_j}) / \tau} })
\end{equation}

Binding all different spaces together results in a hybrid model with $K$ 3D point encoders, $S$ audio encoders, $N$ image encoders, and $M$ text encoders. These encoders originate from different pre-trained models but share the same encoding space after binding. The representations of the resulting ensemble model can be calculated as:
\begin{equation}
\label{eq:ave}
        \Bar{\mathbf{P}} = \sum_{i=1}^K \alpha_i \!\cdot\! \mathbf{P}_i; \ \ \ \Bar{\mathbf{A}} = \sum_{i=1}^S \beta_i \! \cdot\! \mathbf{A}_i; \ \ \ \Bar{\mathbf{V}} = \sum_{i=1}^N \gamma_i \!\cdot\! \mathbf{V}_i;\ \ \ \Bar{\mathbf{T}} = \sum_{i=1}^M \delta_i \!\cdot\! \mathbf{T}_i;
\end{equation}
where the $\Bar{\mathbf{P}}$, $\Bar{\mathbf{A}}$, $\Bar{\mathbf{V}}$, $\Bar{\mathbf{T}}$ are the embeddings of resulting model, and $\mathbf{P}_i$, $\mathbf{A}_i$, $\mathbf{V}_i$, $\mathbf{T}_i$ respectively denote the 3D point, audio, image, and text representations from each corresponding $i$-th encoder. The $\alpha_i$, $\beta_i$, $\gamma_i$, $\delta_i$ are the combining factors for $i$-th encoder of each modality.

\subsection{Weights Routing}
Existing works~\cite{wang2023connecting, wang2023extending, wang2024freebind} manually set the combining factors of encoders from different spaces. While manual settings offer flexibility to customize the resulting space when integrating a few spaces, it is increasingly complex and impractical as more spaces are added. Moreover, hand-designed combining weights also limit the deep integration across various knowledge sources, leading to simple trade-offs rather than comprehensive incorporation of different expertise.

To address this issue, drawing inspiration from the Mixture-of-Expert (MoE) technique in Large Language Models (LLMs), we propose dynamically assigning weights with learnable routers. As shown in Fig.~\ref{fig:model}, each modality contains one router to predict the corresponding combining factors for encoders of this modality, which can be formulated as:
\begin{equation}
\label{eq:router}
\begin{gathered}
    \alpha_1, \dots, \alpha_K = \operatorname{softmax}(\Theta_p(\mathbf{P})); \ \beta_1, \dots, \beta_S = \operatorname{softmax}(\Theta_a(\mathbf{A}))\\ 
    \gamma_1, \dots, \gamma_N = \operatorname{softmax}(\Theta_v(\mathbf{V})); \ \delta_1, \dots, \delta_M = \operatorname{softmax}(\Theta_t(\mathbf{T}))
\end{gathered}
\end{equation}
where $\mathbf{P}$, $\mathbf{A}$, $\mathbf{V}$, $\mathbf{T}$ are the concatenated outputs of each modality, and the $\Theta_p$, $\Theta_a$, $\Theta_v$, $\Theta_t$ denote the router for 3D point, audio, image and language, respectively. To develop effective and robust routers, we utilize the entire retrieved pseudo dataset $\{p, a, v, t\}$ and design two learning objectives:

\textbf{Cross-modal Overall Alignment. } To motivate routers to predict the optimal weights for all modality combinations, we employ contrastive losses overall modality pairs as the first learning target:
\begin{equation}
\label{eq:L_align}
    L_{align} = \operatorname{Info}(\mathbf{\Bar{A}}, \mathbf{\Bar{P}}) + \operatorname{Info}(\mathbf{\Bar{A}}, \mathbf{\Bar{V}}) + \operatorname{Info}(\mathbf{\Bar{A}}, \mathbf{\Bar{T}}) + \operatorname{Info}(\mathbf{\Bar{P}}, \mathbf{\Bar{V}}) + \operatorname{Info}(\mathbf{\Bar{P}}, \mathbf{\Bar{T}}) + \operatorname{Info}(\mathbf{\Bar{V}}, \mathbf{\Bar{T}})
\end{equation}
where $\mathbf{\Bar{P}}$, $\mathbf{\Bar{A}}$, $\mathbf{\Bar{V}}$, $\mathbf{\Bar{T}}$ are defined in Eq.~\ref{eq:ave}. By simply averaging the contrastive losses between all modality pairs, we cultivate balanced routers that achieve comprehensively high-quality cross-modal semantic alignment over all modalities.  

\textbf{Language Representation Decoupling. } Compared to 3D points, audio, and images, which are primarily sampled from the real world, language data is entirely artificial, exhibiting much higher information density and a stronger ideographic tendency. Therefore, textual descriptions of different modalities exhibit significant biases: image captions often describe appearances, audio captions focus on sounding actions, and 3D captions prioritize spatial structures. As a result, text encoders trained to align different modalities demonstrate more specialized expertise than encoders of other modalities.

Considering the significant distribution variance among the different text representations, we introduce an auxiliary learning objective for the language router to disentangle the language representation and improve its generalization. It preserves the discrimination of text embedding space and enhances the semantic alignments with various modalities. Specifically, we drive the language router to identify which modality the input texts are likely describing and to prioritize text encoders that are specialized in the corresponding modality. To this end, we define the loss function as follows:
\begin{equation}
\label{eq:l_dec}
    L_{dec} = -\sum^M_{j=1} [y_j \operatorname{log}(\Theta_t(\mathbf{T})_j) + (1-y_j) \operatorname{log}(1-\Theta_t(\mathbf{T})_j)]
\end{equation}
where the $M$ is the number of text encoders. The texts in the retrieved pseudo-paired dataset are collected from audio-text, image-text, and 3D-text pairs. Correspondingly, the text encoders are also derived from models pre-trained for audio-text, image-text, and 3D-text alignment. In Eq.~\ref{eq:l_dec}, $y_j=1$ if the input text and the $j$-th text encoder are related to the same modality and $y_j=0$ otherwise.

Finally, we linearly combine the above two objectives, and the final loss can be expressed as:
\begin{equation}
\label{eq:loss}
    L = \lambda L_{dec} + L_{align}
\end{equation}

\subsection{Model Configurations}
The projectors $\Psi$ used for aligning spaces are simple two-layer MLPs. Additionally, we employ the mixture-of-projectors strategy following~\cite{wang2024freebind}. The routers $\Theta$ are similarly designed as two-layer MLPs, with an extra sigmoid activation function at the end.

We select 14 pre-trained spaces for binding, which can be grouped into five audio-text (three WavCaps~\cite{mei2023wavcaps}, two LIANO-CLAPs~\cite{laionclap2023}), five image-text (EVA-CLIP-18B~\cite{sun2024eva}, EVA02-CLIP-E~\cite{fang2023eva2} two SigLIPs~\cite{zhai2023sigmoid}, DFN-ViT-H~\cite{fang2023data}), three 3D-image-text (three Uni3Ds~\cite{zhou2023uni3d}) and one audio-image-text (ImageBind~\cite{girdhar2023imagebind}) spaces. After binding all spaces to EVA-CLIP-18B, we construct three configurations of OmniBind by combining different spaces: \textbf{OmniBind-Base}, \textbf{OmniBind-Large}, and \textbf{OmniBind-Full} with 7 billion, 13 billion and 30 billion parameters. The encoder parameters for each modality and the specific spaces used in each variant are detailed in Appendix~\ref{sec:variants}.


\section{Experiment}
\subsection{Implementation}
\textbf{Datasets \& Hyper-parameter. } 
To construct the pseudo-paired data, we collect unpaired 3D point, audio, image, and text data from the training set of existing datasets. For 3D data, we use the 800k 3D point clouds from Objaverse~\cite{deitke2023objaverse}. The audio and image data come from AudioSet~\cite{gemmeke2017audio} and ImageNet~\cite{deng2009imagenet} respectively. The text data sources from three kinds of datasets: 3D-text~\cite{liu2024openshape}, visual-text~\cite{lin2014microsoft, sharma2018conceptual} and audio-text~\cite{kim2019audiocaps, drossos2020clotho} datasets. Based on these unpaired unimodal data, we employ state-of-the-art audio-text (WavCaps~\cite{mei2023wavcaps}), image-text (EVA-CLIP-18B~\cite{sun2024eva}), audio-image (ImageBind~\cite{girdhar2023imagebind}) and 3D-image-text (Uni3D~\cite{zhou2023uni3d}) models to retrieve the pseudo item pairs, as discussed in Sec.~\ref{sec:binding}. The temperature factors in contrastive losses are 0.03, and the $\lambda$ in Eq.~\ref{eq:loss} is 3.


\begin{wraptable}{r}{7.5cm}
\vspace{-1.8\baselineskip}
\caption{Statistic for evaluation tasks and benchmarks.}
\centering
\setlength\tabcolsep{5pt}
\label{tab:eval_stat}
\renewcommand{\arraystretch}{1.2}
\resizebox{7.5cm}{!}{
\begin{tabular}{cc|cr}
\toprule
Task & Modality & Benchmarks & Items \\ \midrule
\multirow{6}{*}{\begin{tabular}[c]{@{}c@{}}Zero-shot\\ Classification\end{tabular}} & \multirow{2}{*}{Audio} & AudioSet~\cite{gemmeke2017audio} & 19,048 \\
 &  & ESC-50~\cite{piczak2015esc} & 400 \\ \cline{2-4} 
 \specialrule{0em}{1.5pt}{1.5pt}
 & Image & ImageNet-1K~\cite{deng2009imagenet} & 50,000 \\ \cline{2-4}  \specialrule{0em}{1.5pt}{1.5pt}
 & \multirow{3}{*}{3D} & Objaverse-LVIS~\cite{deitke2023objaverse} & 46,832 \\
 &  & ScanObjNN~\cite{uy2019revisiting} & 2,890 \\
 &  & ModelNet40~\cite{wu20153d} & 2,468 \\ \midrule
\multirow{7}{*}{\begin{tabular}[c]{@{}c@{}}Cross-modal\\ Retrieval\end{tabular}} & \multirow{2}{*}{Audio-Text} & AudioCaps~\cite{kim2019audiocaps} & 964 \\
 &  & Clotho~\cite{drossos2020clotho} & 1,045 \\ \cline{2-4}  \specialrule{0em}{1.5pt}{1.5pt}
 & \multirow{2}{*}{Audio-Image} & VGG-SS~\cite{chen2021localizing} & 5,158 \\
 &  & FlickrNet~\cite{senocak2018learning} & 5,000 \\ \cline{2-4}  \specialrule{0em}{1.5pt}{1.5pt}
 & \multirow{2}{*}{Image-Text} & COCO~\cite{lin2014microsoft} & 5,000 \\
 &  & Flickr-30K~\cite{young2014image} & 1,000 \\ \cline{2-4}  \specialrule{0em}{1.5pt}{1.5pt}
 & 3D-Image & Objaverse-LVIS~\cite{deitke2023objaverse} & 46,205 \\ \bottomrule
\end{tabular}}
\vspace{-1\baselineskip}
\end{wraptable}

\textbf{Benchmarks \& Baselines. } To comprehensively access the performance of our omni representations, we conduct quantitative experiments across 13 benchmarks covering 7 downstream tasks, as summarized in Tab.~\ref{tab:eval_stat}. In these benchmarks, we compare OmniBinds with three groups of previous multimodal representation models: 1) 3D-image-text models: Uni3D's pre-trained and three fine-tuned variants~\cite{zhou2023uni3d}. 2) audio-image-text models: C-MCR~\cite{wang2023connecting}, LanguageBind~\cite{zhu2023languagebind}, ImageBind~\cite{girdhar2023imagebind}, ImageBind++~\cite{wang2024freebind} and InternVL$_{I\!B}$++~\cite{wang2024freebind}. 3) 3D-audio-image-text models: Ex-MCR~\cite{wang2023extending} and PointBind~\cite{guo2023point}. Moreover, we provide the best results on each benchmark achieved by the 14 source specialist spaces for reference, denoted as "Individual Best".


\subsection{Performance Results}

\textbf{Cross-modal Retrieval. }
As aforementioned, OmniBind aims to provide high-quality semantic alignment between all modality pairs. Therefore, we comprehensively assess the cross-modal retrieval performance across all possible modality pairs. The quantitative results for audio-text, audio-image, image-text, and 3D-image retrieval are presented in Tab.~\ref{tab:retrieval}.

Overall, OmniBind-Full and OmniBind-Large consistently outperform all previous methods across all benchmarks. Some prior approaches demonstrate competitive performance within their specific domains. For instance, ImageBind++ shows similarly strong audio-image alignment, and InternVL$_{I\!B}$++ displays comparable image-text capabilities, but they both fall short in other areas and lack support for 3D input. Compared to existing 3D-audio-image-text models like Ex-MCR and PointBind, all OmniBind variants exhibit substantial and comprehensive advantages across all combinations of modalities. These observations underscore the versatility and superiority of OmniBind.


Moreover, OmniBind-Full achieves similar performance to the "Individual Best", demonstrating that OmniBind successfully inherits and effectively integrates the expertise of various source spaces.
Remarkably, OmniBind achieves even better audio-image and 3D-image alignment, showcasing the exciting cross-space knowledge transfer. By incorporating the high-quality image representations learned from image-text data, the audio-image and 3D-image alignment can also be improved.

\begin{table*}[t]
\caption{Cross-modal retrieval results. Best result is \textbf{bolded}, and second best result is \underline{underlined}. }
\centering
\setlength\tabcolsep{2pt}
\label{tab:retrieval}
\renewcommand{\arraystretch}{1.2}
\resizebox{1.02\textwidth}{!}{
\begin{tabular}{l|cccc|cccc|cccc|cc}
\toprule
 \multirow{3}{*}{Models} & \multicolumn{4}{c|}{Audio-Text} & \multicolumn{4}{c|}{Audio-Image} & \multicolumn{4}{c|}{Image-Text} & \multicolumn{2}{c}{3D-Image} \\
& \multicolumn{2}{c}{AudioCaps} & \multicolumn{2}{c|}{Clotho} & \multicolumn{2}{c}{VGG-SS} & \multicolumn{2}{c|}{FlickrNet} & \multicolumn{2}{c}{COCO} & \multicolumn{2}{c|}{Flickr30K} & \multicolumn{2}{c}{Objaverse} \\
& R@1 & R@5 & R@1 & R@5 & R@1 & R@5 & R@1 & R@5 & R@1 & R@5 & R@1 & R@5 & R@1 & R@5 \\ \midrule
Uni3D & $\times$ & $\times$ & $\times$ & $\times$ & $\times$ & $\times$ & $\times$ & $\times$ &59.51 &	81.45 &	87.85 &	97.55 & 43.57 & 67.61   \\ 
C-MCR & 15.76 &	41.37 &	8.37 &	24.86 &	1.94 &	7.69 &	1.39 &	5.97 &	16.67 &	37.04 &	34.16 &	63.64 & $\times$ & $\times$  \\
LanguageBind &12.42 &	36.70&	11.32 	&31.03 &	2.55 &	9.86 &	1.52 &	6.36 &	53.24 &	76.48 &	82.36 &	96.19  & $\times$ & $\times$ \\
ImageBind & 9.24 &	27.47 &	6.64 &	17.28 &	14.82 &	35.67 &	7.68 &	20.79 &	57.28 &	79.54 &	86.04 &	96.97 & $\times$ & $\times$    \\
ImageBind++ & 29.16 &	62.98 &	13.67 &	33.19 	&\underline{15.48} &	\textbf{39.26} &	\underline{8.01} &	\underline{21.87} &	57.01 &	79.23  &	85.91  &	97.03 & $\times$ & $\times$   \\
InternVL$_{I\!B}$++ & 29.11  &	62.30 &	12.66 &	32.75 &	14.40 &	36.78 &	7.74 &	21.85 &	\underline{61.07}& 	\underline{82.00} &	\textbf{89.30} &	\textbf{98.09} & $\times$ & $\times$  \\
Ex-MCR & 19.07 &	47.05 &	7.01 &	22.04 &	2.13 &	8.13 &	1.57 &	5.94& 	40.24 &	64.78 &	71.89 &	90.55 &2.54 &	8.25  \\
PointBind & 9.24 &	27.47 &	6.64 &	17.28 &	14.82 &	35.67 &	7.68 &	20.79 &	57.28 &	79.54 &	86.04 &	96.97 & 5.86 &	14.59 	 \\ \midrule
OmniBind-Base & 43.61 &	76.02 &	20.94 &	46.77 &	14.11 	&35.74& 	7.67 &	21.65 &	56.94 &80.11 &	85.99 &	97.02  &	34.34 &	58.40   \\
OmniBind-Large & \textbf{47.89} &	\textbf{79.75} &	\underline{23.07} &	\textbf{49.67} &	14.14 &	36.07 &	7.86 &	21.72& 	60.08 &	82.35 &	87.20 &	97.40 &  \underline{46.09} &	\underline{69.11} 	\\
OmniBind-Full & \underline{46.72} &	\underline{79.69} &	\textbf{23.27} &	\underline{49.46} &	\textbf{15.64} &	\underline{38.19} &	\textbf{8.32} &	\textbf{23.49} &	\textbf{62.64} &	\textbf{83.79} 	&\underline{89.13} &\underline{97.82}  &  \textbf{46.55} &	\textbf{69.92} \\
\midrule
\textcolor{lightgray}{Individual Best} & \textcolor{lightgray}{48.22} &	\textcolor{lightgray}{81.15} 	&	\textcolor{lightgray}{23.57} &	\textcolor{lightgray}{49.13} 	&	\textcolor{lightgray}{14.82} &	\textcolor{lightgray}{35.67} &	\textcolor{lightgray}{7.68} &	\textcolor{lightgray}{20.79}	&  \textcolor{lightgray}{63.69} 	&  \textcolor{lightgray}{84.09} &	\textcolor{lightgray}{90.83} &	\textcolor{lightgray}{98.33}  & \textcolor{lightgray}{43.57} & \textcolor{lightgray}{67.61}	\\
 \bottomrule
 \end{tabular}
 }
 \vspace{-1\baselineskip}
\end{table*}

\begin{table*}[t]
\caption{Zero-shot classification results. Uni3D, Uni3D(Objav.), Uni3D(Scan.) and Uni3D(Model.) represent the pre-trained and three fine-tuned version of Uni3D-g, respectively.}
\centering
\setlength\tabcolsep{5pt}
\label{tab:class}
\renewcommand{\arraystretch}{1.2}
\resizebox{1\textwidth}{!}{
\begin{tabular}{l|ccc|cc|cccccc}
\toprule
\multirow{3}{*}{Model} & \multicolumn{3}{c|}{Audio} & \multicolumn{2}{c|}{Image} & \multicolumn{6}{c}{3D} \\ 
 & \multicolumn{1}{c}{AudioSet} & \multicolumn{2}{c|}{ESC-50} & \multicolumn{2}{c|}{ImageNet} & \multicolumn{2}{c}{Objaverse} & \multicolumn{2}{c}{ScanObjectNN} & \multicolumn{2}{c}{ModelNet40} \\
 & mAP & Top1 & Top5 & Top1 & Top5 & Top1 & Top5 & Top1 & Top5 & Top1 & Top5 \\ \midrule
Uni3D  & $\times$ & $\times$ & $\times$ & 80.12 &	95.97  & 53.13 &	81.59 &	{64.12} &	91.63  & 87.56 &	\textbf{{99.27}}  \\
Uni3D(Objav.) & $\times$ & $\times$ & $\times$ &  80.12 &	95.97  & \textbf{54.74} &	\textbf{82.54} &	58.89 &	88.69 &	84.20 & 98.42   \\
Uni3D(Scan.) & $\times$ & $\times$ & $\times$ & 80.12 &	95.97  & 50.99 &	80.00 &	\textbf{65.81} &	{92.70} &	\underline{88.05} &	98.58   \\
Uni3D(Model.) & $\times$ & $\times$ & $\times$ & \underline{80.12} &	\underline{95.97}  & 51.21 &	80.14 &	64.43 &	90.66 &	\textbf{88.05} & \underline{99.11}  \\
C-MCR  & 11.15 &	70.35 &	96.70 &	24.50 &	52.44  & $\times$ & $\times$ & $\times$ & $\times$ & $\times$ & $\times$ \\
LanguageBind  & 18.33 &	\textbf{94.00} &	99.70 &	77.15 &	94.64  &  $\times$ & $\times$ & $\times$ & $\times$ & $\times$ & $\times$ \\
ImageBind  & 13.96 &	67.25 &	87.50 &	76.31 &	94.23  & $\times$ & $\times$ & $\times$ & $\times$ & $\times$ & $\times$ \\
ImageBind++  & 19.69 &	90.30 &	99.30 &	76.01 &	94.36  & $\times$ & $\times$ & $\times$ & $\times$ & $\times$ & $\times$ \\
InternVL$_{I\!B}$++  & 18.93 &	87.75 &	98.75 &	\textbf{81.21} &	\textbf{96.68}  & $\times$ & $\times$ & $\times$ & $\times$ & $\times$ & $\times$  \\
Ex-MCR  & 6.67 &	71.20 &	96.80 &	60.79 &	86.98  & 17.94 &	43.37 &	40.31 &	77.20  & 66.53 &	93.60  \\
PointBind  & 13.96 &	67.25 &	87.50 &	76.13 &	94.22  & 13.83 &	30.34 &	55.05 &	86.89   & 76.18 &	97.04  \\ \midrule
OmniBind-Base  &21.19  &  92.90 &	99.75  &	76.18 &	94.02  & 53.30 &	81.85 &	57.79 &	89.76   & 82.82 &	97.12   \\
OmniBind-Large  & \textbf{25.57}  &	93.25 	&\underline{99.80}  & 78.87 &	95.32  & \underline{53.97} &	\underline{82.90} &	64.67 &	\underline{94.15}   & 86.55 &	99.03    \\
OmniBind-Full & \underline{25.14}  & \underline{93.45} &	\textbf{99.85}  & {79.93} &	{95.86}   &  53.56 	&81.82 & \underline{64.67} &	\textbf{94.36} &87.12 &	99.03  \\ \midrule
\textcolor{lightgray}{Individual Best}  & \textcolor{lightgray}{23.36} & \textcolor{lightgray}{94.05} & \textcolor{lightgray}{99.75} & \textcolor{lightgray}{82.43} & \textcolor{lightgray}{96.73} & \textcolor{lightgray}{54.74} &	\textcolor{lightgray}{82.54} &	\textcolor{lightgray}{65.81} &	\textcolor{lightgray}{92.70}   & \textcolor{lightgray}{88.05} &	\textcolor{lightgray}{99.11}  \\
\bottomrule
\end{tabular}}
\vspace{-1\baselineskip}
\end{table*}

\textbf{Zero-shot Classification. }
To further validate the generalization ability of OmniBind, we conduct zero-shot classification on each modality, and report the results in Tab.~\ref{tab:class}.

For audio and image classification, OmniBind demonstrates overall superiority, even when compared to models excelling in the audio-image-text domains. While LanguageBind performs slightly better than OmniBind on ESC-50, all variants of OmniBind significantly outperform previous methods on the more challenging AudioSet benchmark. This highlights the robustness and generalization of OmniBind. Furthermore, although InternVL$_{I\!B}$++ achieves the highest accuracy on ImageNet, it falls short in audio classification, which further showcases the versatility of OmniBind.

In 3D object classification, the three fine-tuned variants of Uni3D perform exceptionally well on their respective fine-tuned datasets. OmniBind-Base, which leverages only Uni3D(Objav.) as the 3D-image-text source space, achieves performance comparable to Uni3D(Objav.), demonstrating that the binding space effectively inherits knowledge from the source space. Additionally, OmniBind-Large and OmniBind-Full integrate the knowledge from all three fine-tuned versions, performing well across various benchmarks and exhibiting the best overall performance.



\begin{figure}[t]
	\centering
	\includegraphics[width=1\linewidth]{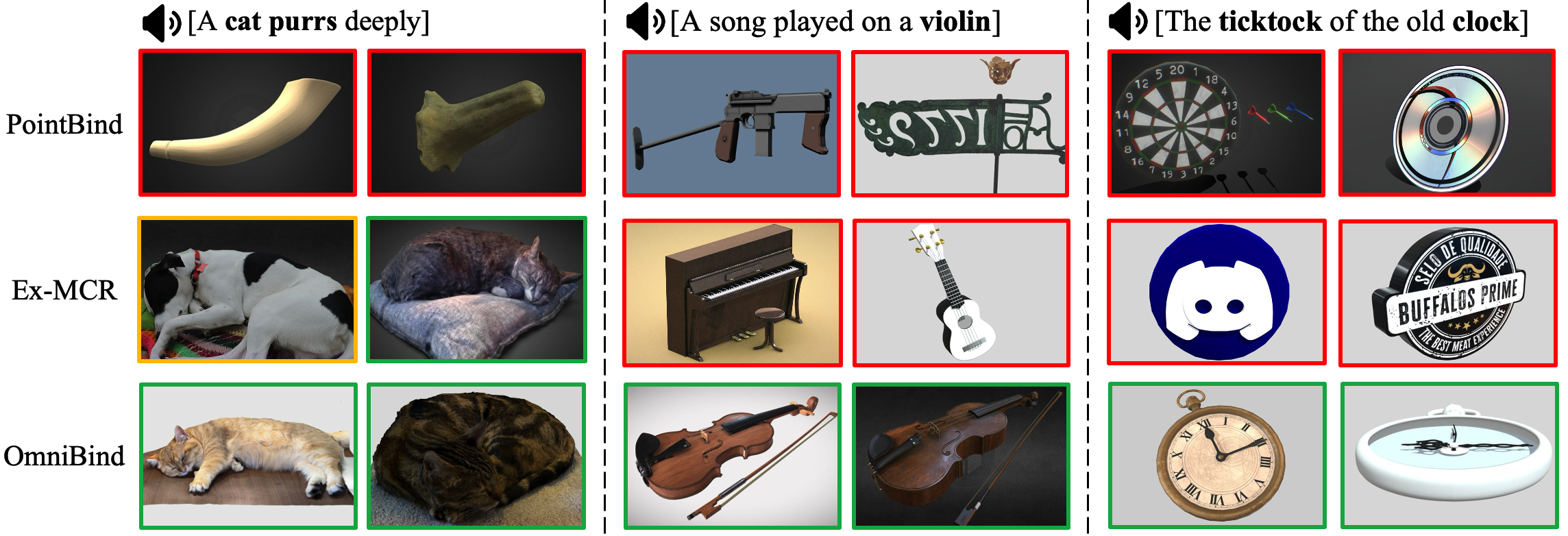}
        \vspace{-1.5\baselineskip}
	\caption{Qualitative comparison of audio to 3D object retrieval. More visualizations of 3D-audio retrieval are provided in the Appendix~\ref{sec:visualization}.}
        \vspace{-0.5\baselineskip}
        \label{fig:audio-3d}
\end{figure}

\begin{figure}[t]
	\centering
	\includegraphics[width=1\linewidth]{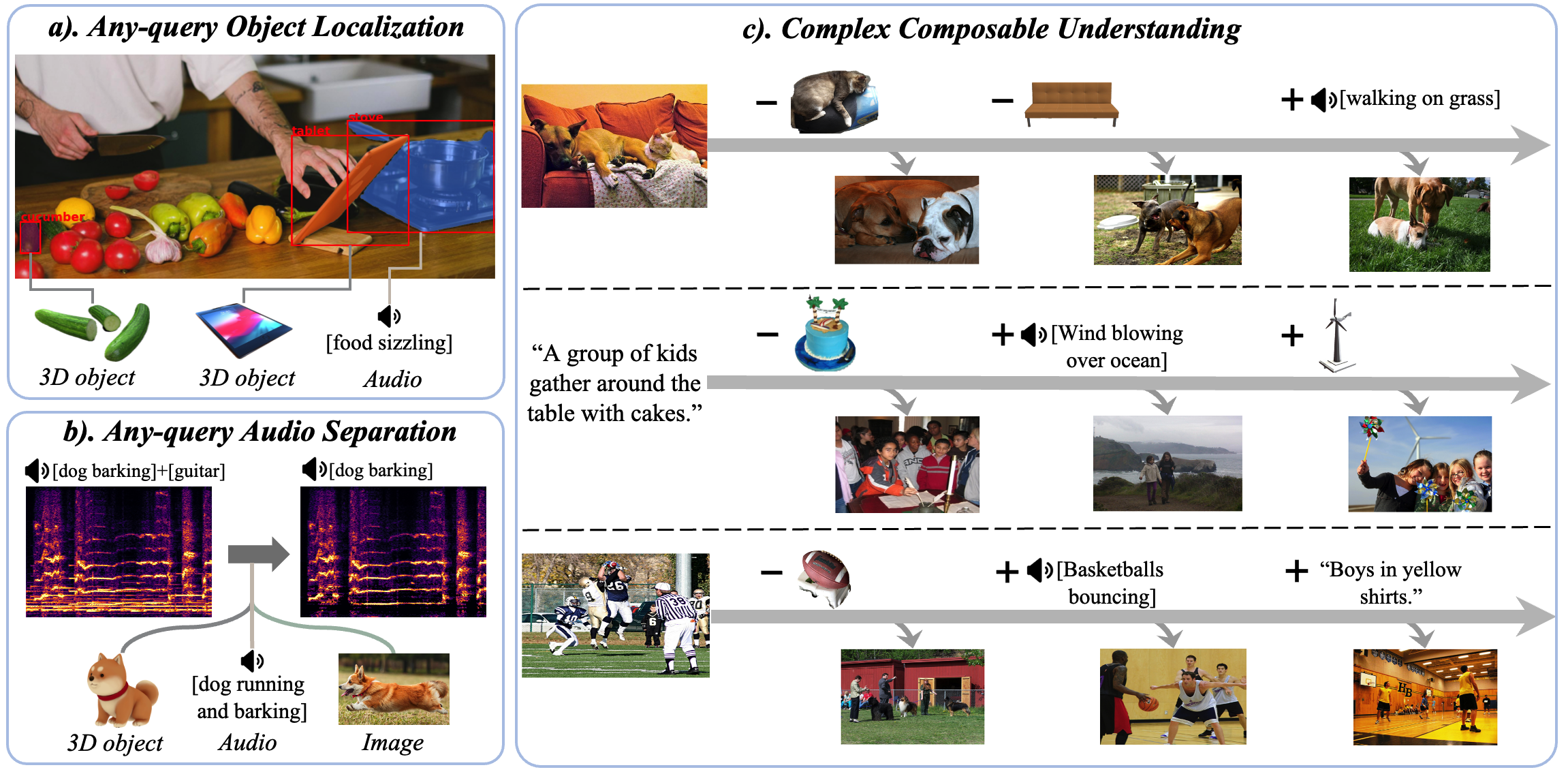}
        \vspace{-1.5\baselineskip}
	\caption{Diverse applications enabled by OmniBind.}
        \vspace{-0.5\baselineskip}
        \label{fig:application}
\end{figure}

\subsection{Applications}
\textbf{Emergent cross-modal alignment. }
OmniBind reveals a surprising emergent alignment between 3D and audio. In Fig.~\ref{fig:audio-3d} and \ref{fig:1}, we randomly sample audios from AudioCaps and retrieve relevant 3D objects in Objaverse. OmniBind shows a deep understanding of each modality and high-quality semantic alignment. For example, it accurately recognizes the audio of "cat purring" and "violin" and finds the corresponding 3D models. In the "clock ticktock" case, while PointBind and Ex-MCR struggle to distinguish disk-shaped 3D models, OmniBind successfully identifies mechanical clocks.

\textbf{Any-query object localization. } OmniBind also demonstrates high-quality fine-grain semantic alignment, enabling any-query object localization. By extracting object proposals using pre-trained object segmentation~\cite{kirillov2023segment} models, we can match these object proposals with queries from any modality within OmniBind's space. As shown in Fig.~\ref{fig:application}-\textit{a}, we can localize objects in images using sounds or 3D models as queries.


\textbf{Any-query audio separation. } We replace the CLIP embeddings in CLIPSep~\cite{dong2022clipsep} with OmniBind and fine-tune it on our pseudo dataset. In Fig.~\ref{fig:application}-\textit{b}, we observe that we can successfully separate audios of a dog barking using queries from images, audio, language, and even 3D models of a dog.

\textbf{Complex composable understanding. } 
Impressively, OmniBind's embedding space not only enables addition and subtraction between arbitrary embeddings but also supports complex serial embedding arithmetic. Three representative cases are provided in Fig.~\ref{fig:application}-\textit{c}. In the first case, starting with a photo of "one cat and one dog on sofa," we sequentially subtract the embeddings of a 3D cat and a 3D sofa to obtain "two dogs on sofa" and "two dogs not on sofa." Adding the sound of walking on grass then results in "two dogs on the grass." Similar exciting capabilities are observed in two other examples.


\textbf{More potentials. } Previous multimodel joint representations~\cite{zhou2023uni3d, girdhar2023imagebind, zhu2023languagebind} enable other applications, like point cloud painting~\cite{zhou2023uni3d}, any-to-any generation~\cite{tang2024any}, and omnimulti modal LLMs~\cite{han2023imagebind, lin2023video}. Given that OmniBind is essentially homologous to the representation models used in previous applications and contains more diverse and generalizable knowledge, it holds great potential for these applications as well. We leave these explorations for future work.

\begin{table*}[t]
\vspace{-1\baselineskip}
\caption{Ablation study of manual weights and the two weight routing objectives: $L_{align}$ and $L_{dec}$. \textit{AT high}, \textit{VT high}, and \textit{PVT high} indicate manually assigning higher weights to audio-text, image-text, and 3D-image-text spaces, respectively. \textit{Mean} means averaging all spaces. The Top1 and R@1 metrics of 3D classification and cross-modal retrieval are reported. The dataset names are abbreviated.}
\centering
\setlength\tabcolsep{3pt}
\label{tab:ablation}
\renewcommand{\arraystretch}{1.2}
\resizebox{1.02\textwidth}{!}{
\begin{tabular}{ccc|ccc|cc|cc|cc|c}
\toprule
\multicolumn{1}{c}{\multirow{2}{*}{Setting}} & \multirow{2}{*}{$L_{align}$} & \multicolumn{1}{l|}{\multirow{2}{*}{$L_{dec}$}} & \multicolumn{3}{c|}{3D classification} & \multicolumn{2}{c|}{Audio-Text} & \multicolumn{2}{c|}{Audio-Image} & \multicolumn{2}{c|}{Image-Text} & \multicolumn{1}{l}{3D-Image} \\
\multicolumn{1}{l}{} &  & \multicolumn{1}{l|}{} & Objav. & Scan. & Model. & \multicolumn{1}{l}{ACaps} & \multicolumn{1}{l|}{Clotho} & \multicolumn{1}{l}{VGGSS} & \multicolumn{1}{l|}{FNet} & \multicolumn{1}{l}{COCO} & \multicolumn{1}{l|}{F30K} & Objav. \\ \midrule
\textit{AT high} &- &- & 49.47  & 59.90  & 85.17  & 44.47 &	21.76 &	14.33 &	7.69 &	55.97 &	84.81  & 46.23  \\
\textit{VT high} & -& -& 53.44  & 64.19  & 86.99  & 33.24 &	17.49 &	13.82 &	7.45 &	61.38 	&\underline{88.57}  & 46.24  \\
\textit{PVT high} &- &- & 53.35  & 63.08  & 87.07  & 34.95 &	18.08 &	13.18 &	7.34 &	59.86 	&87.84  & \textbf{48.18}  \\ 
\textit{Mean} & - & -& 52.29 & 62.46  & 86.47  & 40.31 & 20.33 &	13.40 &	7.35 &	59.61 &	87.61   & 44.61  \\ \midrule
\multirow{3}{*}{\begin{tabular}[c]{@{}c@{}}Weight\\ Routing\end{tabular}} & $\times$ & \checkmark & \textbf{53.70} &	\textbf{64.81}   &\textbf{87.40}  &  \textbf{48.48} &	\textbf{24.31} &	13.93 &	7.35 &	\underline{61.86} &	88.27 &  44.61 \\
 & \checkmark & $\times$ & 52.93 &	63.98  & 85.74 & 43.71 &	20.44 	&\underline{15.62} 	&\textbf{8.35} 	&59.43& 	87.64  & \underline{46.73} \\
 & \checkmark & \checkmark & \underline{53.56} &	\underline{64.67}  & \underline{87.12}  & \underline{46.72} &	\underline{23.27} &	\textbf{15.64} &	\underline{8.32} &	\textbf{62.64} &	\textbf{89.13}  & {46.55}  \\ \bottomrule
\end{tabular}}
\vspace{-1\baselineskip}
\end{table*}

\subsection{Ablation Study}

\textbf{Performance improvement with more spaces. }
The integrated spaces and parameters of our three variants: OmniBind-Base, OmniBind-Large, and OmniBind-Full, sequentially increase. From Tab.~\ref{tab:retrieval} and \ref{tab:class}, we observe consistent overall performance improvements with the growth of the model scale, further proving the effectiveness of "scaling up" representations model via binding more spaces.

\textbf{Trade-offs in manually assigned weights. } Tab.~\ref{tab:ablation} shows the performance of four types of manually assigned weights: \textit{AT high}, \textit{VT high}, \textit{PVT high} and \textit{Mean}.
These four variants reveal trade-off phenomena among different areas of expertise. \textit{AT high} performs well in the audio-text domain but is less effective in 3D classification and image-text tasks. Conversely, \textit{VT high} (\textit{PVT high}) obtains better image-text (3D-image) performance at the expense of audio-text alignment. \textit{Mean} exhibits a more balanced performance but lacks specific expertise. Moreover, using weight routing showcases a comprehensive advantage over these manual weight variants. This observation further proves the routers can effectively alleviate interference between knowledge of different sources, leading to overall improvements in performance rather than simple trade-offs.

\textbf{Effectiveness of $L_{align} \ \& \ L_{dec}$ for learning routers. }
We provide the ablation experiments for the two learning objectives in Tab.~\ref{tab:ablation}. By comparing the \textit{Mean} and variants using each objective separately, we conclude that $L_{align}$ brings noticeable improvements across all possible modality combinations, while $L_{dec}$ specifically and significantly enhances language-related alignment. Combining the two objectives yields the best overall performance by complementing each other's strengths.

\textbf{Improved discrimination with $L_{dec}.$ }
To explore the effect of language representation decoupling on discrimination, we extract language embeddings from all AudioCaps and COCO captions. We find that employing $L_{dec}$ reduces the cosine similarity between all the possible language embedding pairs from 0.0828 to 0.0517. As discussed in~\cite{liang2022mind}, lower cosine similarity between irrelevant pairs indicates the embeddings "occupy" more space in the hypersphere. Therefore, employing $L_{dec}$ indeed enhances the discrimination of language representations, leading to better overall performance.



\section{Conclusion}

In this work, we present OmniBind, large-scale omni multimodal representation models ranging in scale from 7 billion to 30 billion parameters. Our core contributions include successfully scaling up multimodal representation by binding many existing spaces and designing a weight routing strategy to mitigate interference between knowledge of different sources. By combining 14 pre-trained spaces, OmniBind demonstrates impressive multimodal performance across extensive benchmarks and shows vast potential for further growth and diverse applications.

\bibliographystyle{plain}
\bibliography{main}


\newpage
\appendix

\section{Limitations and Future Works}
Although OmniBind is already the largest multimodal representation model, demonstrating remarkable versatility and generalization, it currently utilizes only 14 existing spaces and 4 modalities: 3D, audio, images, and text. It remains to be explored whether or when this binding space-based "scaling up" will reach saturation. Additionally, this work only presents three basic applications as inspiration. Investigating whether using higher-quality and larger-scale omni multimodal representations can enable new capabilities in downstream applications would be a promising direction.

\section{Statistics of OmniBind Variants}
\label{sec:variants}
The detailed source spaces used to build OmniBind-Base, OmniBind-Large, and OmniBind-Full are presented in Tables \ref{tab:base}, \ref{tab:large}, and \ref{tab:full}, respectively. Generally speaking, OmniBind-Base comprises three components: three WavCaps, EVA02-CLIP-E, Uni3D (Objaverse), and ImageBind. The primary difference between OmniBind-Large and OmniBind-Full is the inclusion of EVA-CLIP-18B in OmniBind-Full. Additionally, we provide statistics on the number of parameters for OmniBind-Base, OmniBind-Large, and OmniBind-Full in Table \ref{tab:stat}.

\begin{table*}[h]
\caption{Source spaces for building OmniBind-Base.}
\centering
\setlength\tabcolsep{5pt}
\label{tab:base}
\renewcommand{\arraystretch}{1.2}
\resizebox{0.8\textwidth}{!}{
\begin{tabular}{cc}
\toprule
\multicolumn{2}{c}{OmniBind-Base} \\ \midrule
Audio-Text & WavCaps~\cite{mei2023wavcaps}, WavCaps (Clotho)~\cite{mei2023wavcaps}, WavCaps (AudioCaps)~\cite{mei2023wavcaps} \\ \midrule
Image-Text & EVA02-CLIP-E~\cite{fang2023eva2} \\ \midrule
3D-Image-Text & \multicolumn{1}{c}{Uni3D (Objav.)~\cite{zhou2023uni3d}} \\ \midrule
Audio-Image-Text & ImageBind~\cite{girdhar2023imagebind} \\ \bottomrule
\end{tabular}}
\end{table*}

\begin{table*}[h]
\caption{Source spaces for building OmniBind-Large.}
\centering
\setlength\tabcolsep{5pt}
\label{tab:large}
\renewcommand{\arraystretch}{1.2}
\resizebox{0.8\textwidth}{!}{
\begin{tabular}{cc}
\toprule
\multicolumn{2}{c}{OmniBind-Large} \\ \midrule
Audio-Text & \begin{tabular}[c]{@{}c@{}}WavCaps~\cite{mei2023wavcaps}, WavCaps (Clotho)~\cite{mei2023wavcaps}, WavCaps (AudioCaps)~\cite{mei2023wavcaps},\\  LAION-CLAP (general)~\cite{laionclap2023}, LAION-CLAP (music)~\cite{laionclap2023}\end{tabular} \\ \midrule
Image-Text & EVA02-CLIP-E~\cite{fang2023eva2}, DFN-ViT-H~\cite{fang2023data}, SigLIP-Large~\cite{zhai2023sigmoid}, SigLIP-so400M~\cite{zhai2023sigmoid} \\ \midrule
3D-Image-Text & \multicolumn{1}{c}{Uni3D (Objav.)~\cite{zhou2023uni3d}, Uni3D (Scan.)~\cite{zhou2023uni3d}, Uni3D (Model.)~\cite{zhou2023uni3d}} \\ \midrule
Audio-Image-Text & \multicolumn{1}{c}{ImageBind~\cite{girdhar2023imagebind}} \\ \bottomrule
\end{tabular}}
\end{table*}

\begin{table*}[h]
\caption{Source spaces for building OmniBind-Full.}
\centering
\setlength\tabcolsep{5pt}
\label{tab:full}
\renewcommand{\arraystretch}{1.2}
\resizebox{0.8\textwidth}{!}{
\begin{tabular}{cc}
\toprule
\multicolumn{2}{c}{OmniBind-Full} \\ \midrule
Audio-Text & \begin{tabular}[c]{@{}c@{}}WavCaps~\cite{mei2023wavcaps}, WavCaps (Clotho)~\cite{mei2023wavcaps}, WavCaps (AudioCaps)~\cite{mei2023wavcaps},\\  LAION-CLAP (general)~\cite{laionclap2023}, LAION-CLAP (music)~\cite{laionclap2023}\end{tabular} \\\midrule
Image-Text & \begin{tabular}[c]{@{}c@{}}EVA-CLIP-18B~\cite{sun2024eva}, EVA02-CLIP-E~\cite{fang2023eva2}\\  DFN-ViT-H~\cite{fang2023data}, SigLIP-Large~\cite{zhai2023sigmoid}, SigLIP-so400M~\cite{zhai2023sigmoid} \end{tabular} \\ \midrule
3D-Image-Text & \multicolumn{1}{c}{Uni3D (Objav.)~\cite{zhou2023uni3d}, Uni3D (Scan.)~\cite{zhou2023uni3d}, Uni3D (Model.)~\cite{zhou2023uni3d}} \\\midrule
Audio-Image-Text & \multicolumn{1}{c}{ImageBind~\cite{girdhar2023imagebind}} \\ \bottomrule
\vspace{-1\baselineskip}
\end{tabular}}
\end{table*}

\begin{table*}[t]
\vspace{-1\baselineskip}
\caption{Statistics about parameter number of three OmniBind variants.}
\centering
\setlength\tabcolsep{3pt}
\label{tab:stat}
\renewcommand{\arraystretch}{1.2}
\resizebox{0.9\textwidth}{!}{
\begin{tabular}{l|cccccc}
\toprule
\multirow{2}{*}{Variants} & \multicolumn{6}{c}{Parameter Number} \\
 & 3D Encoder & Audio Encoder & Image Encoder & Text Encoder & Projector & Total \\ \midrule
OmniBind-Base & 1.0B & 99M & 4.6B & 1.3B & 224M & 7.2B \\
OmniBind-Large & 3.0B & 329M & 6.0B & 2.5B & 431M & 12.3B \\
OmniBind-Full & 3.0B & 329M & 23.6B & 3.2B & 431M & 30.6B \\ \bottomrule
\end{tabular}}
\end{table*}

\section{More Visualizations}
\label{sec:visualization}
To further qualitatively assess the capabilities of OmniBind, we provide additional visualizations. Each sample is manually evaluated and categorized: correct samples are marked in green, incorrect ones in red, and partially correct ones in orange.

\begin{figure}[t]
\centering
\includegraphics[width=\linewidth]{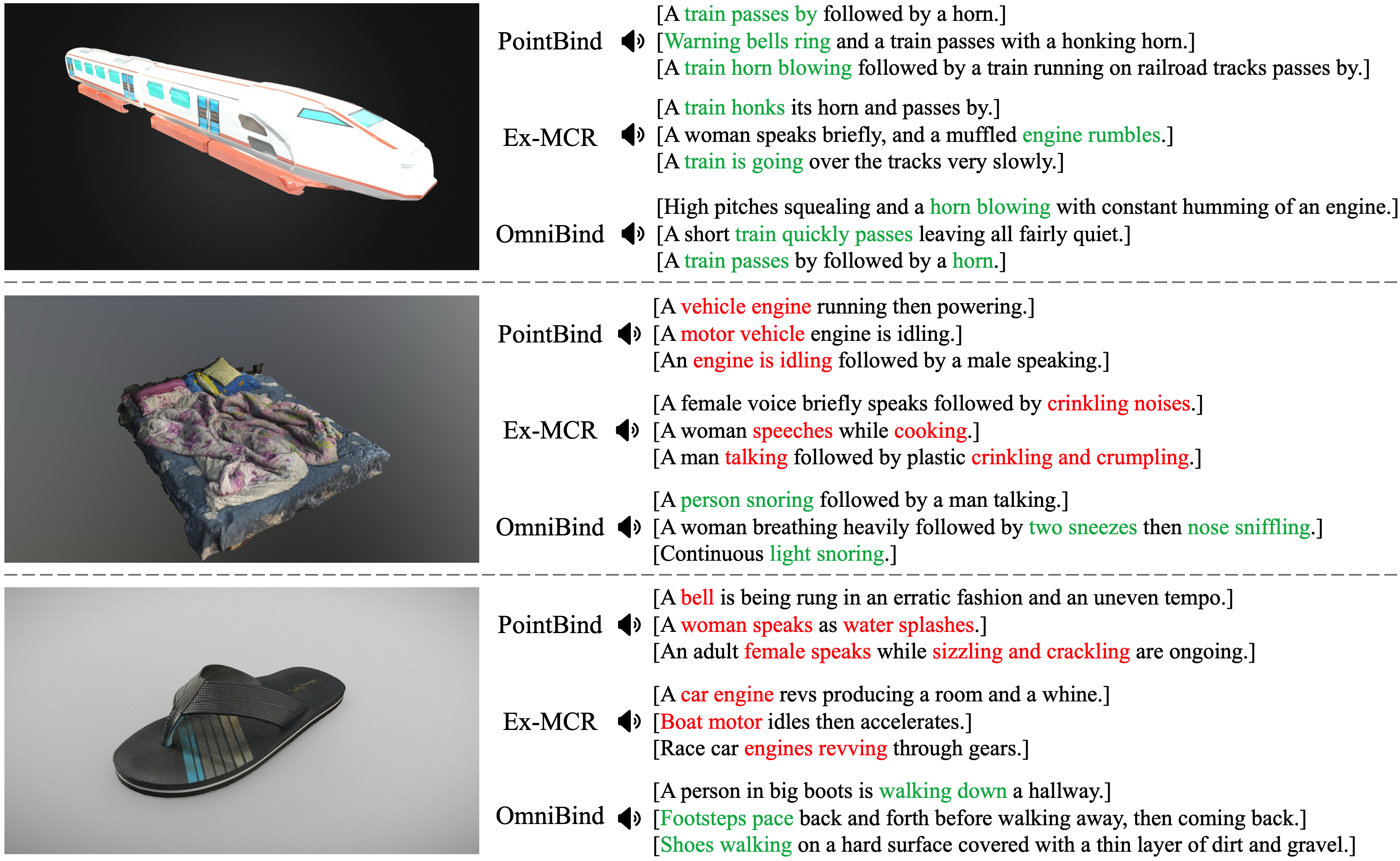}
\caption{More visualization of 3D-to-Audio retrieval.}
\label{fig:3d2audio}
\end{figure}

In Fig.~\ref{fig:3d2audio}, we randomly select three 3D objects from Objaverse and retrieve relevant audio clips from AudioCaps. The captions for the retrieved audio clips are provided. OmniBind demonstrates more accurate retrieval results compared to previous methods (PointBind and Ex-MCR).

\begin{figure}[h]
\centering
\includegraphics[width=0.85\linewidth]{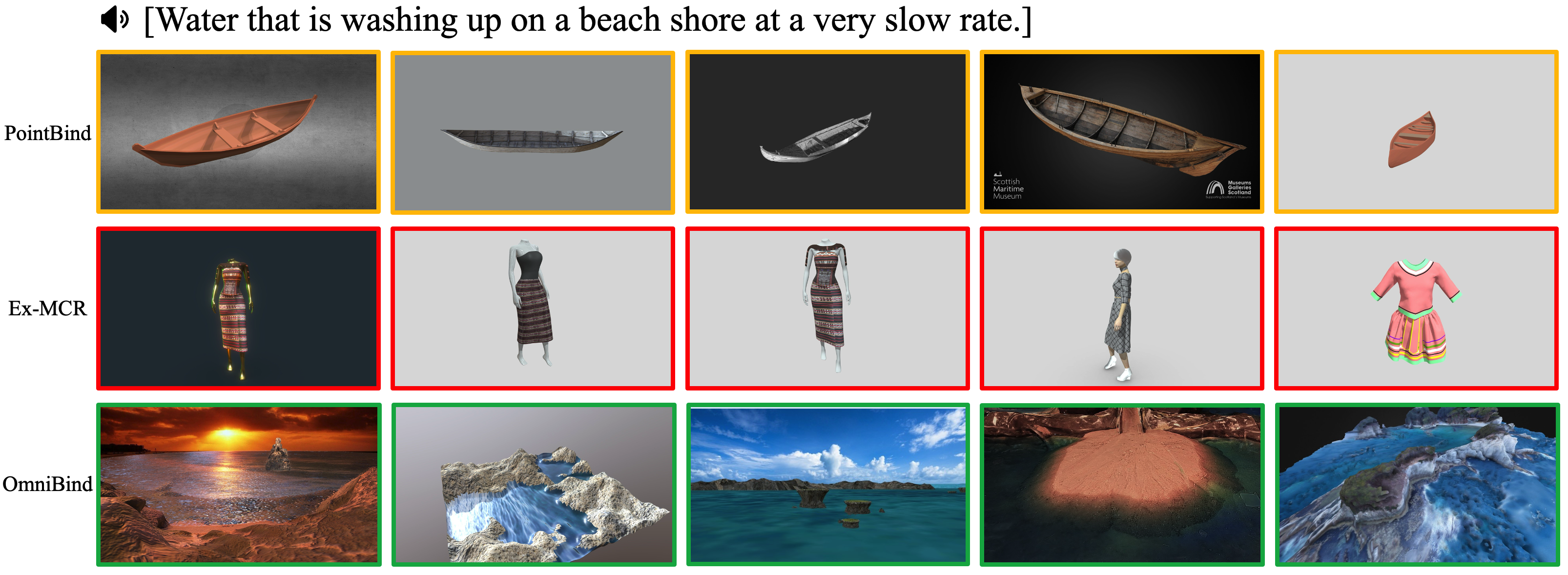} \\
\vspace{0.2\baselineskip} 
\includegraphics[width=0.85\linewidth]{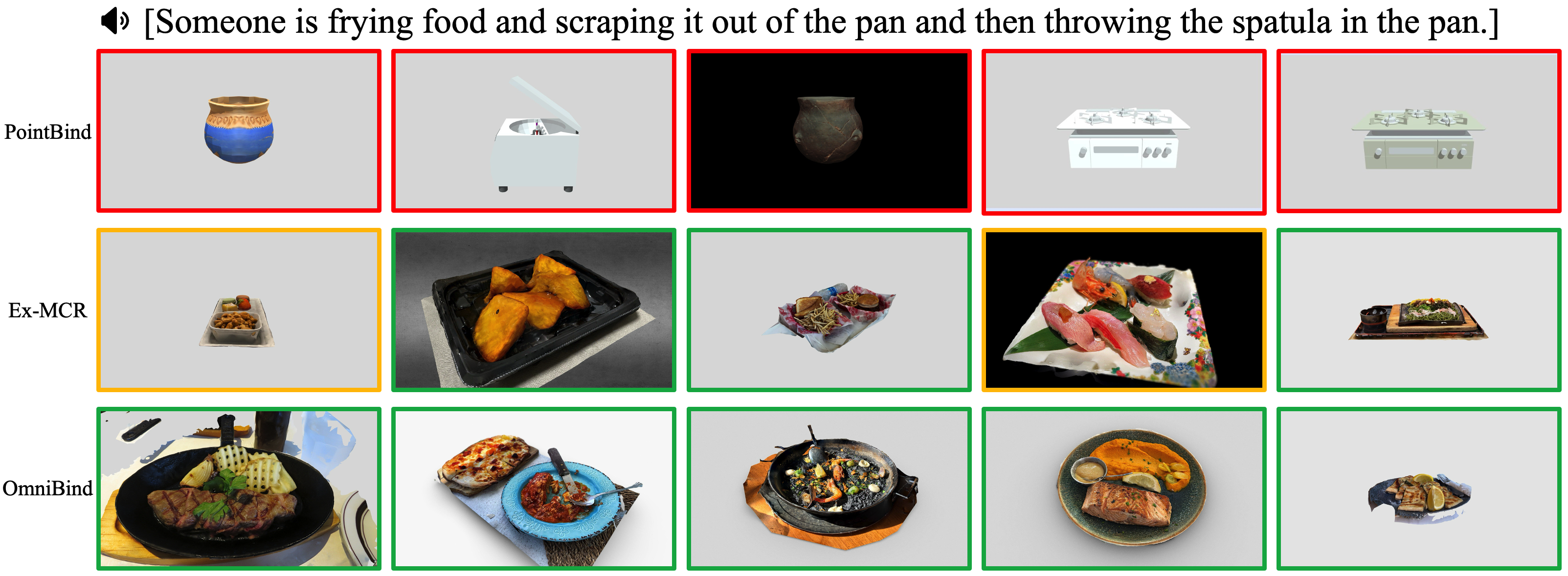} \\
\vspace{0.2\baselineskip} 
\includegraphics[width=0.85\linewidth]{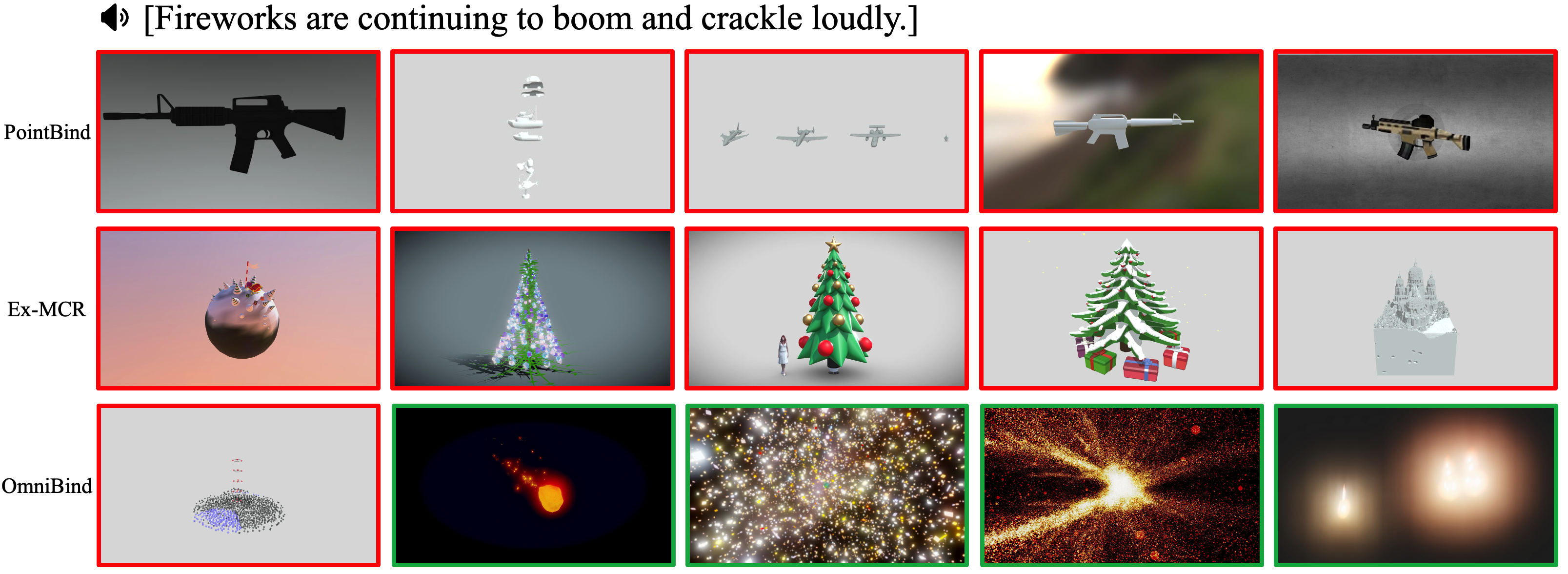} \\
\vspace{0.2\baselineskip} 
\includegraphics[width=0.85\linewidth]{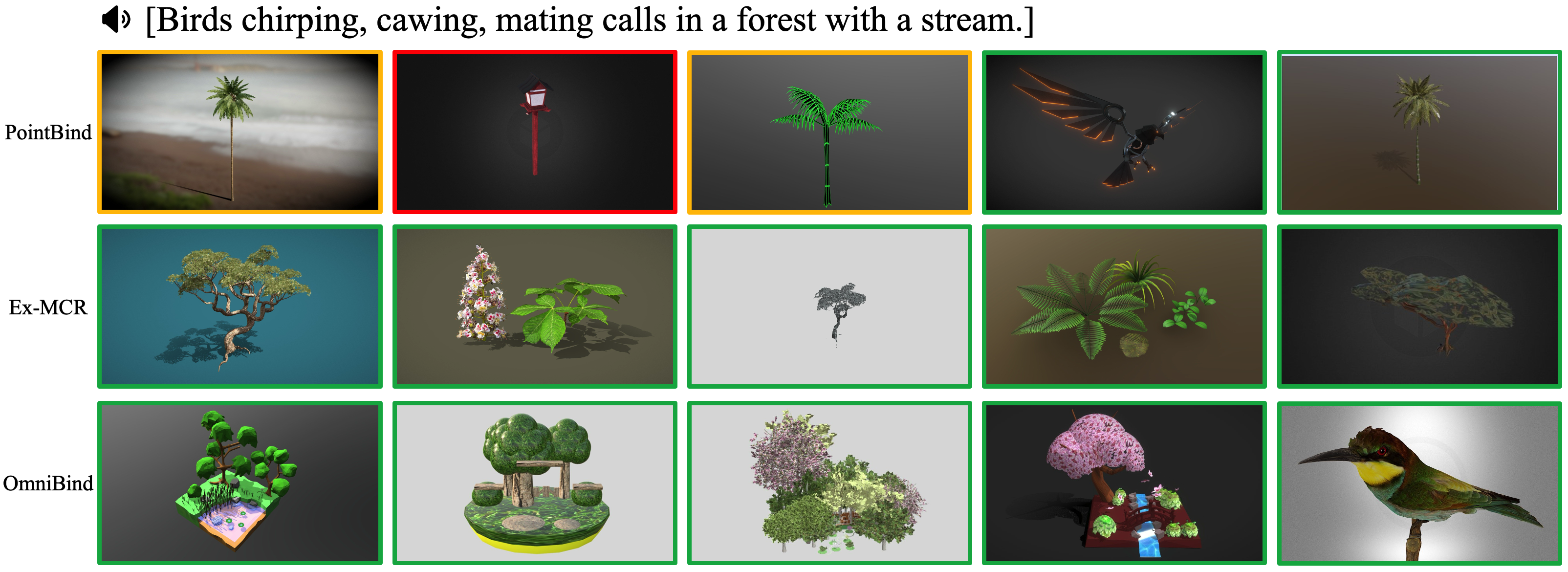}
\vspace{0.2\baselineskip} 
\includegraphics[width=0.85\linewidth]{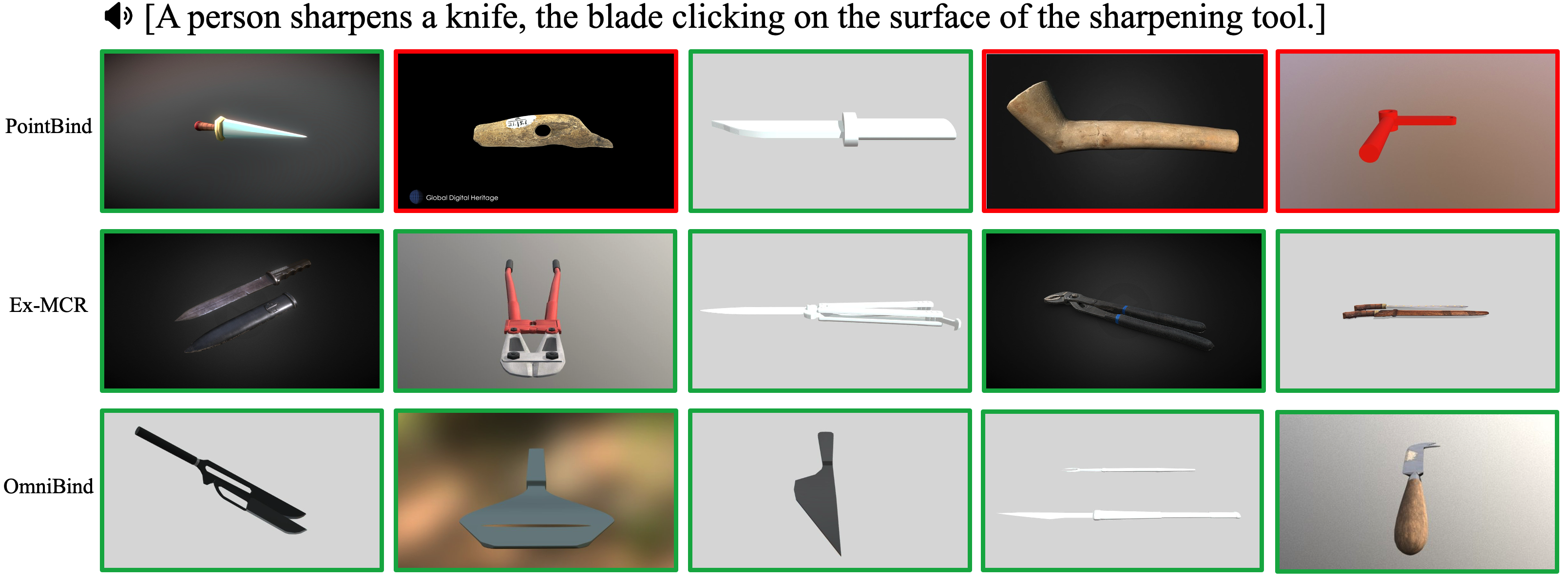} 
\caption{More visualization of Audio-to-3D retrieval.}
\label{fig:1}
\end{figure}

Furthermore, we present additional cases of audio to 3D object retrieval in Figs.~\ref{fig:1}. In each instance, the 3D objects retrieved by OmniBind exhibit significantly better semantic alignment with the audio query than those retrieved by Ex-MCR and PointBind.

\section{Social Impact}
OmniBind is a method for learning large-scale multimodal representation models by aligning and reorganizing knowledge from existing pre-trained models. Therefore, the capability of OmniBind is mainly inherited from the source pre-trained models. Implementing additional safety detection and filtering processes for these pre-trained models can effectively reduce the potential for misuse and mitigate negative social impacts of OmniBind.


\end{document}